\documentclass[lettersize,journal]{IEEEtran}
\usepackage{subcaption}
\usepackage{caption}
\usepackage{booktabs}
\usepackage[T1]{fontenc}
\usepackage{lmodern}  
\usepackage{amsmath,amsfonts}
\usepackage{algorithmic}
\usepackage{algorithm}
\usepackage{array}
\usepackage{textcomp}
\usepackage{stfloats}
\usepackage{url}
\usepackage{verbatim}
\usepackage{graphicx}
\usepackage{cite}
\begin{document}

\title{GADPN: Graph Adaptive Denoising and Perturbation Networks via Singular Value Decomposition}

\author{Hao Deng, Bo Liu, ~\IEEEmembership{Member,~IEEE,}
}



\maketitle

\begin{abstract}
While Graph Neural Networks (GNNs) excel on graph-structured data, their performance is fundamentally limited by the quality of the observed graph, which often contains noise, missing links, or structural properties misaligned with GNNs' underlying assumptions. To address this, graph structure learning aims to infer a more optimal topology. Existing methods, however, often incur high computational costs due to complex generative models and iterative joint optimization, limiting their practical utility. 

In this paper, we propose GADPN, a simple yet effective graph structure learning framework that adaptively refines graph topology via low-rank denoising and generalized structural perturbation. Our approach makes two key contributions: (1) we introduce Bayesian optimization to adaptively determine the optimal denoising strength, tailoring the process to each graph's homophily level; and (2) we extend the structural perturbation method to arbitrary graphs via Singular Value Decomposition (SVD), overcoming its original limitation to symmetric structures. Extensive experiments on benchmark datasets demonstrate that GADPN not only achieves state-of-the-art performance but does so with significantly improved efficiency. It shows particularly strong gains on challenging disassortative graphs, validating its ability to robustly learn enhanced graph structures across diverse network types.
\end{abstract}


\section{Introduction}

\IEEEPARstart{G}{raphs} are ubiquitous in the real world, such as biology networks, social networks and citation networks.
As graphs have become the fundamental data structure in countless applications, 
the demand for learning effective graph representations to enable downstream task performance has never been more pressing.
In recent years,
Graph Neural Networks (GNNs) have become a research hotspot in the field of graph representation learning \cite{10.5555/3294771.3294869,DBLP:conf/iclr/KipfW17,wu2019simplifying,bo2021beyond}.
These GNNs have achieved state-of-the-art performance in many tasks such as node classification 
\cite{xu2018powerful,xu2018representation,abu2019mixhop},
graph classification \cite{gao2019graph,zhang2018end},
and recommender systems \cite{wang2019neural,ying2018graph}.

Although existing GNNs have achieved remarkable success in various graph-based tasks,their performance heavily relies on the quality of input graph structures. To justify their effectiveness, most GNNs assume that the observed graphs are complete and accurate. However, this assumption is usually not satisfied since real-world graphs may be error-prone and often suffer from incompleteness, inaccuracies, and misalignment with GNN inductive biases. For instance, in molecular graphs, traditional laboratory experimental error \cite{10.1145/3442381.3449952} is usually unavoidable. In social networks, implicit social relationships between people often lead to missing links in graphs. Moreover, since most GNNs prefer the low-frequency information than high-frequency information \cite{bo2021beyond,li2018deeper}, even accurate graphs may violate the performance of GNNs when they contain much high-frequency information. This can well explain why most GNNs excel on homophily graphs but struggle with disassortative ones\cite{newman2002assortative}. 

To address the shortcomings of GNNs on real‑world noisy graph data, graph structure learning methods aim to infer or refine graph topologies so as to enhance GNN performance. However, such approaches often encounter several challenges, the foremost being the high computational overhead introduced by multi‑stage iterative optimization. For instance, GEN \cite{10.1145/3442381.3449952} proposes a Bayesian‑inspired graph estimation framework: its structural model employs the Stochastic Block Model (SBM) \cite{holland1983stochastic}, a classic generative model that partitions nodes into blocks and assigns connection probabilities based on block memberships; while SBM itself introduces estimation complexity, GEN further integrates, for the first time, multi‑order neighborhood information into posterior calculation; and the whole framework alternates between updating the graph estimate and tuning the GNN parameters. Similarly, Geom‑GCN \cite{Pei2020Geom-GCN:} introduces a geometry‑based aggregation scheme that first embeds nodes into a continuous latent space to construct structural neighborhoods, then designs a bi‑level aggregator to capture long‑range dependencies. While these methods are theoretically appealing and deliver notable performance gains, they significantly increase computational burden due to the additional embedding steps, multi‑order relation modeling, and iterative inference procedures. In contrast, conventional GNNs such as GCN and GAT remain computationally efficient, yet their reliance on the raw graph structure limits their effectiveness in highly noisy or disassortative scenarios. Therefore, balancing the expressiveness gained through structure learning with manageable computational cost remains a critical challenge for the practical deployment of such advanced graph learning methods.

However, effectively learning an optimal graph structure for GNNs while maintaining computational efficiency remains an open challenge. Moreover, two primary obstacles must be addressed: (1) Generalization across different graph types – a robust method should perform well adaptively on both assortative and disassortative graphs, which exhibit fundamentally different connectivity patterns under the homophily principle; and (2) Mitigation of overfitting – estimating graph structure introduces additional model complexity, which can easily lead to overfitting, especially when labeled data are scarce.

Among existing approaches, two lines of research are particularly relevant: (1) link prediction via structural perturbation, and (2) graph enhancement via matrix factorization. The structural perturbation method (SPM) proposed by Lu et al.\cite{lu2015toward} estimates missing links by perturbing the observed network, grounded in the insight that links are predictable if their removal does not significantly alter the network’s structural features. This offers a principled perspective for inferring robust topology from noisy observations. However, SPM is inherently limited to symmetric adjacency matrices of undirected graphs, making it difficult to extend to directed or attributed graphs. On the other hand, LightGCL\cite{cai2023lightgcl} employs singular value decomposition (SVD) for low-rank graph approximation, effectively capturing global collaborative signals and improving computational efficiency in recommendation systems. While this demonstrates the strength of low-rank approximation in preserving essential information and suppressing noise, LightGCL does not incorporate a structural perturbation mechanism, lacks a theoretical framework for explicit graph refinement through perturbation, and often relies on empirical selection of the approximation rank without theoretical guidance.

In summary, these two lines of work offer valuable insights—SPM through perturbation-based inference and LightGCL through low-rank condensation—yet each has notable limitations: SPM is restricted by symmetry and undirected assumptions, while LightGCL does not integrate perturbation ideas and lacks theoretical grounding for rank selection. Therefore, a key direction for advancing graph structure learning lies in effectively combining the robustness of perturbation-based inference with the global efficiency of low-rank approximation, and extending such a framework to broader graph types and tasks.

\textbf{Our Contributions:} In this paper, we propose Graph Adaptive Denoising and Perturbation Networks(GADPN), a novel graph structure learning framework that \textbf{unifies and extends} both paradigms. Our key innovations are:
\begin{itemize}
    \item \textbf{Generalization of Structural Perturbation:} We extend the perturbation method of \cite{lu2015toward} to general (possibly asymmetric) graphs using SVD, overcoming its original limitation to symmetric matrices.
    \item \textbf{Enhanced Low-Rank Approximation:} Beyond the SVD-based low-rank approximation in \cite{cai2023lightgcl}, we incorporate Bayesian-optimized adaptive denoising and structural perturbation to prevent overfitting and improve robustness.
    \item \textbf{Theoretical and Empirical Validation:} We provide a comprehensive formulation of the combined approach and demonstrate its effectiveness across diverse homophily levels.
\end{itemize}

The remainder of this paper is organized into four sections. Section \ref{secRelatedWorks} reviews the related literature. Section \ref{secMethod} is devoted to the preliminaries and the proposed methodology. Experimental results and analyses are provided in Section \ref{secExperiment}. Finally, we conclude the paper in Section \ref{secConclusion}.

\section{RELATED WORK}\label{secRelatedWorks}

\subsection{Spectral Graph Neural Networks}
Spectral GNNs define convolution operations by leveraging the spectral domain of graph signals. The pioneering work of Spectral CNN \cite{DBLP:journals/corr/BrunaZSL13} performed convolution in the Fourier domain using learnable diagonal filters, but its dependency on full eigendecomposition of the graph Laplacian incurred high computational cost. ChebNet \cite{defferrard2016convolutional} improved efficiency by approximating spectral filters with Chebyshev polynomials, enabling localized filtering. GCN \cite{DBLP:conf/iclr/KipfW17} further simplified this framework to a first-order approximation with added self-loops, enhancing interpretability and scalability. SGC \cite{wu2019simplifying} demonstrated that a simple linear model applied to precomputed higher-order graph diffusion could achieve competitive performance, underscoring the critical role of graph structure over intricate neural architectures.

\subsection{Spatial Graph Neural Networks}
Spatial GNNs design aggregation functions that operate directly on a node's local neighborhood. GraphSAGE \cite{10.5555/3294771.3294869} introduced a general inductive framework by sampling and aggregating features from a node's neighbors, enabling generalization to unseen nodes. GAT \cite{velickovic2018graph} enhanced aggregation by employing attention mechanisms to assign learnable, feature-dependent weights to neighbors. Building on this, AGNN \cite{k.2018attentionbased} applied attention to learn the importance of both neighbors and edges. Other approaches integrate broader structural insights: APPNP \cite{gasteiger2018combining} derived an improved propagation scheme based on personalized PageRank, effectively combining local and global information. FAGCN \cite{bo2021beyond} proposed a self-gating mechanism to adaptively blend different signals during message passing. To address the static attention limitation of GAT, GATv2 \cite{brody2022how} introduced a dynamic attention mechanism that is provably more expressive and robust to feature perturbations.

\subsection{Graph Structure Learning}
Graph Structure Learning (GSL) methods aim to infer or refine the underlying graph topology from potentially noisy observational data to enhance downstream task performance. Early approaches framed this as a joint optimization problem. Bayesian GCNN \cite{zhang2019bayesian} treated the graph as a random variable from a learnable ensemble. LDS \cite{franceschi2019learning} formulated it as a bilevel program to learn the graph and GNN parameters simultaneously. To improve robustness, Pro-GNN \cite{10.1145/3394486.3403049} reconstructed graphs by enforcing structural priors like sparsity and low rank. GEN \cite{10.1145/3442381.3449952} specifically generated community-structured graphs via Bayesian inference, integrating multi-hop neighborhood information. Alternatively, Geom-GCN \cite{Pei2020Geom-GCN:} redefined local neighborhoods by projecting nodes into a latent geometric space. A common challenge among these methods is balancing expressiveness with computational overhead, often involving iterative optimization or complex generative models.

\subsection{Link Prediction and Structural Perturbation}
Link prediction aims to infer missing or future connections in networks, a task deeply connected to understanding network structural regularities. Heuristic methods like Common Neighbors (CN) \cite{cn} and the Adamic-Adar index \cite{Adamic} leverage local neighborhood overlap, with the latter weighting rare neighbors more heavily. SimRank \cite{simrank} measures similarity recursively based on the principle that nodes are similar if connected to similar nodes. More pertinent to our work is the Structural Perturbation Method (SPM) \cite{lu2015toward}, which estimates missing links by analyzing the stability of a network's eigenvectors under small perturbations. This approach provides a principled, global perspective on link predictability based on structural consistency, moving beyond local heuristics. However, SPM is inherently limited to symmetric, undirected graphs and operates primarily as a preprocessing step rather than being integrated into an end-to-end learning framework.

\noindent
In summary, while GSL methods seek to learn better graphs for GNNs, and link prediction offers tools like SPM to infer structure, a gap remains in efficiently and adaptively integrating perturbation-based robustness with low-rank global approximation for general graph types. Our work bridges this gap by drawing inspiration from both lines of research.

\begin{figure*}[htbp]
  \includegraphics[width=\textwidth]{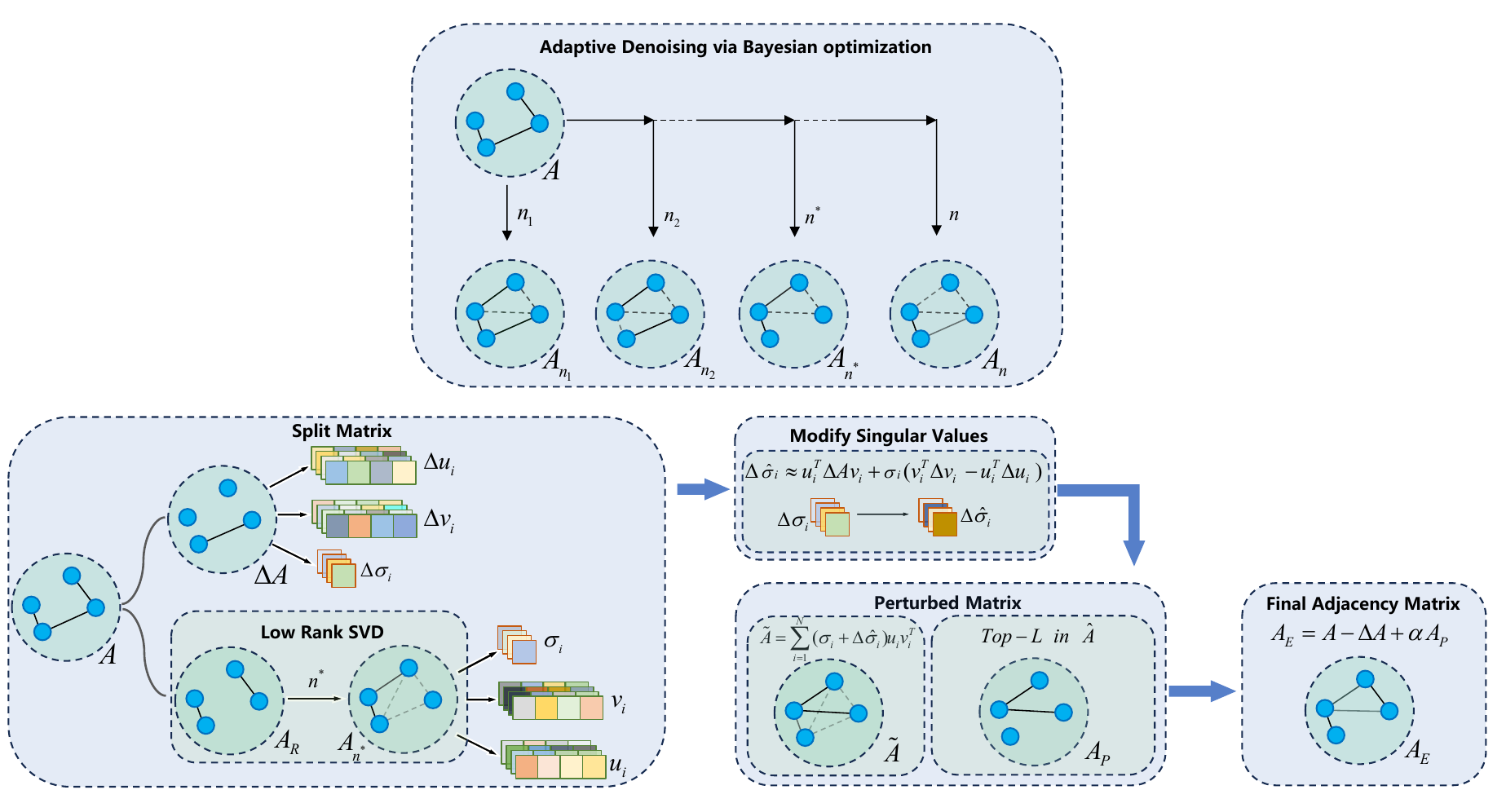}
  \caption{The framework of GADPN.}
  \label{fig:framework}
\end{figure*}

\section{THE PROPOSED METHOD: GADPN}
\label{secMethod}

This section presents \textbf{GADPN}, a novel and efficient graph structure learning framework for GNNs. GADPN synergistically integrates Singular Value Decomposition (SVD) for low-rank denoising with a generalized structural perturbation mechanism to refine graph topology. Our main contributions are threefold: (1) We propose a \textit{generalized structural perturbation} method based on SVD that extends the original structural perturbation method (SPM) \cite{lu2015toward} from symmetric to arbitrary (including directed) graphs. (2) We introduce \textit{Bayesian optimization} to adaptively determine the optimal rank for SVD-based denoising, overcoming the heuristic or fixed-rank limitations in prior work \cite{cai2023lightgcl}. (3) We integrate these two components into a unified, lightweight framework that significantly enhances GNN performance on both assortative and disassortative graphs.

\subsection{Preliminaries}
\label{subsec:preliminaries}
We consider an attributed graph $\mathcal{G} = (\mathcal{V}, \mathcal{E}, \mathbf{X})$, where $\mathcal{V}=\{v_1, v_2, \dots, v_N\}$ is the set of $N$ nodes, $\mathcal{E} \subseteq \mathcal{V} \times \mathcal{V}$ is the set of edges, and $\mathbf{X} \in \mathbb{R}^{N \times d}$ is the node feature matrix with $d$-dimensional features. The graph structure is represented by the adjacency matrix $\mathbf{A} \in \{0, 1\}^{N \times N}$, where $\mathbf{A}_{ij}=1$ if $(v_i, v_j) \in \mathcal{E}$, and $\mathbf{A}_{ij}=0$ otherwise. We focus on the semi-supervised node classification task, where labels $\mathbf{Y}_L = \{y_1, \dots, y_L\}$ are provided only for a small subset of nodes $\mathcal{V}_L \subseteq \mathcal{V}$. As for the experiment, we use GCN and GraphSAGE as the backbone structure. In GCN, each layer is defined as:
\begin{equation}
\mathbf{H}^{(l+1)} = \sigma \left( \tilde{\mathbf{D}}^{-\frac{1}{2}} \tilde{\mathbf{A}} \tilde{\mathbf{D}}^{-\frac{1}{2}} \mathbf{H}^{(l)} \mathbf{W}^{(l)} \right),
\end{equation}
where $\tilde{\mathbf{A}}=\mathbf{A}+\mathbf{I}$ with $\mathbf{A}$ being the adjacency matrix. $\tilde{D}$ is the diagonal degree matrix, and $\sigma(\cdot)$ is a nonlinear activation. In Graph\-SAGE, the node representations are learned through a different approach:
\begin{equation}
\mathbf{h}_{v}^{l}=\sigma\left(\left(\mathbf{h}_{v}^{l-1}\right) \mathbf{W}_{1}^{l}+\left(\operatorname{mean}_{u \in \mathcal{N}(v)} \mathbf{h}_{u}^{l-1}\right) \mathbf{W}_{2}^{l}\right),
\end{equation}
where $\operatorname{mean}_{u \in \mathcal{N}(v)}\mathbf{h}_{u}^{l-1}$ computes the average embedding of the neighboring nodes of $v$.

\subsection{Overview and Motivation}
\label{subsec:overview}
Most GNNs operate under the assumption that the observed graph $\mathcal{G}$ is a complete and accurate representation of the underlying system. However, real-world graphs are often \textit{imperfect}: they may contain spurious edges, lack critical connections, or possess a structure misaligned with the homophily or heterophily principles that GNNs implicitly exploit. Directly applying GNNs on such raw graphs can lead to suboptimal performance.

Existing graph enhancement approaches face several limitations: (1) Generative model-based methods like GEN are computationally expensive and involve complex multi-stage optimization. (2) The structural perturbation method (SPM) provides a principled framework for link prediction but is inherently limited to symmetric (undirected) graphs. (3) Low-rank approximation methods like LightGCL using SVD show promise for noise reduction but lack a principled way to determine the optimal rank and do not incorporate explicit perturbation for robustness.

To address these gaps, we propose GADPN, whose overall architecture is illustrated in Fig. \ref{fig:framework}. GADPN consists of two core modules applied sequentially: \textbf{Adaptive Denoising} and \textbf{Generalized Structural Perturbation}. First, the adaptive denoising module applies rank-adaptive SVD to the original adjacency matrix, where the optimal rank is determined via Bayesian optimization to filter out noise while preserving dominant structural patterns. Second, the generalized structural perturbation module applies a perturbation-and-recovery mechanism to the denoised graph, enhancing its robustness and generalization capacity. The final refined adjacency matrix $\mathbf{A}_E$ can seamlessly replace the original $\mathbf{A}$ as input to any mainstream GNN backbone (e.g., GCN, GAT, GraphSAGE).

\subsection{Adaptive Denoising via Bayesian-Optimized SVD}
\label{subsec:adaptive_denoise}
Real-world graphs contain noise of varying nature and intensity. A fixed denoising strategy is suboptimal across diverse datasets. Our goal is to perform \textit{adaptive} denoising by preserving the most informative components of the graph spectrum via SVD.

Given the adjacency matrix $\mathbf{A}$, its singular value decomposition is:
\begin{equation}
    \mathbf{A} = \mathbf{\mathbf{U}} \boldsymbol{\Sigma} \mathbf{V}^{\mathsf{T}},
\end{equation}
where $\mathbf{U}, \mathbf{V} \in \mathbb{R}^{N \times N}$ are orthogonal matrices whose columns are respectively the left- and right-singular vectors of $\mathbf{A}$, and $\boldsymbol{\Sigma} = \operatorname{diag}(\sigma_1, \sigma_2, \dots, \sigma_N)$ is a diagonal matrix of singular values of $\mathbf{A}$ ordered such that $\sigma_1 \geq \sigma_2 \geq \dots \geq \sigma_N \geq 0$. Larger singular values typically correspond to more dominant structural patterns, while smaller ones are deemed as interfering noises. In real world applications, it is usually computationally expensive to perform an exact SVD on a large graph, so we follow \cite{cai2023lightgcl} to adopt the randomized SVD proposed in \cite{halko2011finding} for a low-rank approximation. The fundamental idea of randomized SVD is first to approximate the range of the input matrix using a low-rank orthonormal matrix, and the carry out SVD on this low-rank approximation, thus significantly alleviates the computational burden. After obtaining $\hat{\mathbf{U}}_n \in \mathbb{R}^{N\times n}$, $\hat{\mathbf{S}}_n \in \mathbb{R}^{n\times n}$,
 $\hat{\mathbf{V}}_n \in \mathbb{R}^{n\times N}$ as the approximated versions of $\mathbf{U}_n$,
 $\boldsymbol{\Sigma}_n$, $\mathbf{V}_{n}^{\mathsf{T}}$, where $n$ is the required rank for the decomposed matrices, we can reconstruct the approximated adjacency matrix as
 \begin{equation}
  {\mathbf{A}}_{n} = \hat{\mathbf{U}}_n \hat{\boldsymbol{\Sigma}}_{n} \hat{\mathbf{V}}_{n}^{\mathsf{T}}.
 \end{equation}
Then a denoising procedure is carried out by retaining only the top $k<n$ singular values and corresponding :vectors so that at last we obtain a rank-$k$ approximation of $\mathbf{A}$:
\begin{equation}
    \mathbf{A}_k = \mathbf{U}_k \boldsymbol{\Sigma}_k \mathbf{V}_k^{\mathsf{T}},
\end{equation}
where $\mathbf{U}_k, \mathbf{V}_k \in \mathbb{R}^{N \times k}$ and $\boldsymbol{\Sigma}_k \in \mathbb{R}^{k \times k}$.

In this procedure, the challenge becomes an effecient determination of the optimal rank $k$. A small $k$ may oversmooth the graph, while a large one may retain excessive noise. We frame this as a black-box optimization problem:
\begin{equation}
    k^* = \arg\max_{k \in \mathcal{K}} \mathcal{P}(k),
\end{equation}
where $\mathcal{K} = \{1, 2, \ldots, n\}$ is the search space, and $\mathcal{P}(k)$ is a performance metric (e.g., validation accuracy) obtained from a fast pre-training routine using the rank-$k$ approximation $\mathbf{A}_k$. Since evaluating $\mathcal{P}(k)$ is computationally expensive and its functional form is unknown, we employ \textbf{Bayesian Optimization (BO)}\cite{BayesianOptimization,frazierTutorialBayesianOptimization2018}.

BO models the unknown function $\mathcal{P}(k)$ as a Gaussian Process (GP):
\begin{equation}
    \mathcal{P}(k) \sim \mathcal{GP}(m(k), \kappa(k, k')),
\end{equation}
with a mean function $m(k)$ (often set to zero) and a covariance kernel $\kappa(k, k')$. Here we adopt the widely used Matérn kernel\cite{rasmussenGaussianProcessesMachine2008} with smoothness parameter $\nu=\frac{3}{2}$:
\begin{equation}
\kappa(k, k') = \sigma^2 \left(1 + \frac{\sqrt{3}|k-k'|}{l}\right)
\exp\left(-\frac{\sqrt{3}|k-k'|}{l}\right),
\end{equation}
where $\sigma^2$ and $l>0$ are the variance parameter and length scale parameter, respectively.

Given a set of observations $\mathcal{D} = \{(k_i, \mathcal{P}(k_i))\}$, BO uses GP to predict the mean $\mu(k)$ and variance $\sigma^2(k)$ of $\mathcal{P}(k)$ at a new point $k$. To select the next point $k_{\text{next}}$ for evaluation, we maximize the Expected Improvement (EI) acquisition function:
\begin{equation}
    \alpha_{\text{EI}}(k) = \mathbb{E}[\max(\mathcal{P}(k) - \mathcal{P}^+, 0)],
\end{equation}
where $\mathcal{P}^+$ is the best observed performance. This process iteratively balances exploration and exploitation, converging to the optimal rank $k^*$ with relatively few evaluations. The final denoised adjacency matrix from this module is denoted as $\hat{\mathbf{A}} = \mathbf{A}_{k^*}$.

\subsection{Generalized Structural Perturbation via SVD}
\label{subsec:perturbation}
While denoising improves the signal-to-noise ratio, the resulting graph structure may still be prone to overfitting. Inspired by the link between structural predictability and stability \cite{lu2015toward} and the known vulnerability of GNNs to adversarial edge perturbations \cite{pmlr-v80-dai18b}, we introduce a \textbf{Generalized Structural Perturbation} module. The core idea is to deliberately perturb a small portion of the graph, then recover and reinforce it, thereby promoting the learning of a more robust and generalizable representation that is resilient to structural noise.

\subsubsection{From Eigenvalue Perturbation to Singular Value Perturbation}
The classic Structural Perturbation Method (SPM) \cite{lu2015toward} is rooted in the first-order perturbation theory for the eigenvalues of a \textit{symmetric} matrix. For a symmetric adjacency matrix $\mathbf{A}_R$ and a small perturbation matrix $\Delta\mathbf{A}$, the shift in the $i$-th eigenvalue $\lambda_i$ is approximated by:
\begin{equation}
    \Delta \lambda_i \approx \mathbf{x}_i^\mathsf{T} \Delta\mathbf{A} \, \mathbf{x}_i,
    \label{eq:original_spm}
\end{equation}
where $\mathbf{x}_i$ is the eigenvector corresponding to $\lambda_i$. This elegant formulation underpins SPM's success in link prediction for undirected graphs. However, its direct application is limited to symmetric matrices, restricting its use to undirected graphs.

To extend this powerful concept to arbitrary directed and undirected graphs, we generalize the perturbation theory to the singular value decomposition (SVD). SVD provides a universal matrix decomposition applicable to any real matrix, making it the ideal tool for this generalization.

\subsubsection{Derivation of Generalized Singular Value Perturbation}
We randomly remove a small fraction $p$ of the original adjacency matrix $\mathbf{A}$ to form a perturbation set $\Delta \mathcal{E}$, with its corresponding adjacency matrix as $\Delta \mathbf{A}$. The residual graph is then ${\mathbf{A}}_R = {\mathbf{A}} - \Delta {\mathbf{A}}$, with its corresponding denoised adjacency matrix as $\hat{\mathbf{A}}_R$.

The SVD of the residual matrix is:
\begin{equation}
    \hat{\mathbf{A}}_R = \sum_{i=1}^{N} \sigma_i \mathbf{u}_i \mathbf{v}_i^{\mathsf{T}},
\end{equation}
where $\sigma_i$ are the singular values, and $\mathbf{u}_i$, $\mathbf{v}_i$ are the corresponding left and right singular vectors, satisfying $\|\mathbf{u}_i\| = \|\mathbf{v}_i\| = 1$ and $\mathbf{u}_i^\mathsf{T}\mathbf{u}_j = \mathbf{v}_i^\mathsf{T}\mathbf{v}_j = \delta_{ij}$.

If we treat $\Delta \hat{\mathbf{A}}$ as a small perturbation to $\hat{\mathbf{A}}_R$, then the perturbed system satisfies:
\begin{equation}
    (\hat{\mathbf{A}}_R + \Delta \hat{\mathbf{A}})(\mathbf{v}_i + \Delta \mathbf{v}_i) = (\sigma_i + \Delta \sigma_i)(\mathbf{u}_i + \Delta \mathbf{u}_i).
    \label{eq:perturbed_system}
\end{equation}
Expanding Eq.~\eqref{eq:perturbed_system} and using the original relation $\hat{\mathbf{A}}_R \mathbf{v}_i = \sigma_i \mathbf{u}_i$, we obtain:
\begin{equation}
    \hat{\mathbf{A}}_R \Delta \mathbf{v}_i + \Delta \hat{\mathbf{A}} \mathbf{v}_i + \Delta \hat{\mathbf{A}} \Delta \mathbf{v}_i = \sigma_i \Delta \mathbf{u}_i + \Delta \sigma_i \mathbf{u}_i + \Delta \sigma_i \Delta \mathbf{u}_i.
\end{equation}
Neglecting the second-order terms $\Delta \hat{\mathbf{A}} \Delta \mathbf{v}_i$ and $\Delta \sigma_i \Delta \mathbf{u}_i$, the equation simplifies to:
\begin{equation}
    \hat{\mathbf{A}}_R \Delta \mathbf{v}_i + \Delta \hat{\mathbf{A}} \mathbf{v}_i \approx \sigma_i \Delta \mathbf{u}_i + \Delta \sigma_i \mathbf{u}_i.
    \label{eq:first_order}
\end{equation}

To isolate $\Delta \sigma_i$, we left-multiply both sides of Eq.~\eqref{eq:first_order} by $\mathbf{u}_i^\mathsf{T}$. Noting that $\mathbf{u}_i^\mathsf{T} \hat{\mathbf{A}}_R = \sigma_i \mathbf{v}_i^\mathsf{T}$ (from $\hat{\mathbf{A}}_R^\mathsf{T} \mathbf{u}_i = \sigma_i \mathbf{v}_i$), we get:
\begin{align*}
    \mathbf{u}_i^\mathsf{T} \hat{\mathbf{A}}_R \Delta \mathbf{v}_i + \mathbf{u}_i^\mathsf{T} \Delta \hat{\mathbf{A}} \mathbf{v}_i &\approx \sigma_i \mathbf{u}_i^\mathsf{T} \Delta \mathbf{u}_i + \Delta \sigma_i \mathbf{u}_i^\mathsf{T} \mathbf{u}_i 
\end{align*}
Therefore,
\begin{align}
   \Delta \sigma_i &\approx \mathbf{u}_i^\mathsf{T} \Delta \hat{\mathbf{A}} \mathbf{v}_i+\sigma_i \left(\mathbf{v}_i^\mathsf{T} \Delta \mathbf{v}_i-\mathbf{u}_i^\mathsf{T} \Delta \mathbf{u}_i\right) . \label{eq:singular_value_perturbation}
\end{align}
If we further assume the orthonormality of the singular vectors is approximately preserved under the small perturbation (i.e., $\mathbf{u}_i^\mathsf{T} \Delta \mathbf{u}_i \approx 0$ and $\mathbf{v}_i^\mathsf{T} \Delta \mathbf{v}_i \approx 0$), then the terms involving $\Delta \mathbf{u}_i$ and $\Delta \mathbf{v}_i$ become negligible. Consequently, Eq.~\eqref{eq:singular_value_perturbation} can be further simplified as:
\begin{equation}
    \Delta \sigma_i \approx \mathbf{u}_i^\mathsf{T} \Delta \hat{\mathbf{A}} \, \mathbf{v}_i.
    \label{eq:singular_value_perturbation_simplified}
\end{equation}

\textbf{Connection to the Original SPM:} In the special case of an undirected graph, $\hat{\mathbf{A}}_R$ is symmetric, and its SVD reduces to its eigen-decomposition with $\mathbf{u}_i = \mathbf{v}_i = \mathbf{x}_i$ and $\sigma_i = |\lambda_i|$. Substituting this into either Eq.~\eqref{eq:singular_value_perturbation} or Eq.~\eqref{eq:singular_value_perturbation_simplified} recovers the original eigenvalue perturbation formula $\Delta \lambda_i \approx \mathbf{x}_i^\mathsf{T} \Delta \hat{\mathbf{A}} \, \mathbf{x}_i$ for positive eigenvalues. Therefore, our generalized singular value perturbation theory strictly contains the original SPM as a special case, establishing its theoretical foundation for arbitrary graph structures.

\subsubsection{Perturbed Matrix Reconstruction and Edge Recovery}
Using the perturbed singular values $\tilde{\sigma}_i = \sigma_i + \Delta \sigma_i$ from Eq.~\eqref{eq:singular_value_perturbation} and the original, unperturbed singular vectors $(\mathbf{u}_i, \mathbf{v}_i)$, we reconstruct an estimated adjacency matrix:
\begin{equation}
    \widetilde{\mathbf{A}} = \sum_{i=1}^{N} \tilde{\sigma}_i \, \mathbf{u}_i \mathbf{v}_i^{\mathsf{T}}.
    \label{eq:perturbed_matrix_recon}
\end{equation}
The values in $\widetilde{\mathbf{A}}$ represent a refined estimation of connection strengths. Crucially, for the edges that were deliberately removed ($\Delta \mathcal{E}$), their corresponding entries in $\widetilde{\mathbf{A}}$ provide a \textit{recovery score} quantifying their likelihood of being true positives within the stable structure of the graph.

The recovery process aims to identify and reintroduce plausible edges that are consistent with the graph's stable structure. To this end, we rank all non-observed edges---i.e., the set $\mathcal{U} \setminus {\mathbf{A}}_R$, which includes the perturbed edges $\Delta \mathcal{E}$---in descending order according to their estimated scores in $\widetilde{\mathbf{A}}$. From this ranked list, we select the top-$P$ edges to form a candidate set $\mathcal{E}^P$, where the number of recovered edges $P$ is determined by:
\begin{equation}
    P = \max\left( \lfloor (p + q)|\mathcal{E}| \rfloor, 0 \right).
    \label{eq:recovery_number}
\end{equation}
Here, $p$ is the base perturbation ratio, and $q$ is a flexible adjustment parameter. The parameter $q$ can be positive, zero, or negative, thereby dynamically controlling whether we recover more, exactly the same, or fewer edges than were initially removed. This mechanism adds a refined and adaptive element to graph reconstruction. The adjacency matrix corresponding to the edge set $\mathcal{E}^P$ is denoted as $\mathbf{A}_P$.

\subsection{Integrated Framework and Final Refined Graph}
\label{subsec:integration}
The two core modules—Adaptive Denoising and Generalized Structural Perturbation—are integrated sequentially into the GADPN pipeline. The final enhanced adjacency matrix $\mathbf{A}_E$, which is passed to the downstream GNN, is synthesized as follows:
\begin{equation}
    \mathbf{A}_E = {\mathbf{A}} - \Delta\mathbf{A} + \alpha \mathbf{A}_P.
    \label{eq:final_adjacency}
\end{equation}
This formulation has an intuitive interpretation: it starts with the original graph ${\mathbf{A}}$, injects robustness by temporarily \textit{deleting} a set of edges $(-\Delta\mathbf{A})$, and finally \textit{recovers and reinforces} plausible connections, potentially including previously missing ones, via the term $+\alpha\mathbf{A}_P$. The hyperparameter $\alpha \in [0, 1]$ controls the influence of the recovered edges \cite{li2018adaptive}.

The complete GADPN workflow is formalized in Algorithm \ref{alg:gadpn}. The computational complexity is dominated by the randomized SVD, which is $\mathcal{O}(N k^2 + |\mathcal{E}|k)$, and the lightweight Bayesian optimization loop, ensuring the framework remains efficient and scalable for practical applications.

\begin{algorithm}[t]
\caption{GADPN: Graph Adaptive Denoising and Perturbation Networks}
\label{alg:gadpn}
\renewcommand{\algorithmicrequire}{\textbf{Input:}}
\renewcommand{\algorithmicensure}{\textbf{Output:}}
\begin{algorithmic}[1]
\REQUIRE Adjacency matrix $\mathbf{A}$, feature matrix $\mathbf{X}$, BO search space $\mathcal{K}$, perturbation ratio $p$, increment $q$, weight $\alpha$.
\ENSURE Enhanced adjacency matrix $\mathbf{A}_E$.
\vspace{0.1cm}
\STATE \textbf{Initialization:}
\STATE $\Delta\mathcal{E} \leftarrow$ Sample $p \cdot |\mathcal{E}|$ edges randomly from $\mathcal{E}$
\STATE $\mathbf{A}_R \leftarrow \mathbf{A} - \Delta\mathbf{A}$ \hfill \COMMENT{\textit{Residual graph construction}}
\vspace{0.2cm}

\STATE \textbf{Phase 1: Adaptive Denoising (via BO)}
\STATE Initialize Gaussian Process $\mathcal{GP}$ over $\mathcal{K}$
\STATE $k^* \leftarrow \text{BayesianOptimization}(\mathbf{A}_R, \mathcal{GP}, \mathcal{K})$ \hfill \COMMENT{\textit{Find optimal rank}}
\STATE $\hat{\mathbf{A}}_R \leftarrow \operatorname{ApproxSVD}(\mathbf{A}_R, k^*)$ \hfill \COMMENT{\textit{Low-rank approximation}}
\vspace{0.2cm}

\STATE \textbf{Phase 2: Generalized Structural Perturbation}
\STATE $[\mathbf{U}, \mathbf{\Sigma}, \mathbf{V}] \leftarrow \operatorname{SVD}(\hat{\mathbf{A}}_R)$
\FOR{each singular value $\sigma_i$ in $\mathbf{\Sigma}$}
    \STATE $\Delta \sigma_i \approx \mathbf{u}_i^\mathsf{T} \Delta \hat{\mathbf{A}} \, \mathbf{v}_i$ \hfill \COMMENT{\textit{Eq. \ref{eq:singular_value_perturbation_simplified}}}
    \STATE $\tilde{\sigma}_i \leftarrow \sigma_i + \Delta \sigma_i$
\ENDFOR
\STATE $\widetilde{\mathbf{A}} \leftarrow \sum_{i} \tilde{\sigma}_i \mathbf{u}_i \mathbf{v}_i^{\mathsf{T}}$ \hfill \COMMENT{\textit{Perturbed matrix reconstruction}}
\vspace{0.1cm}
\STATE \textit{// Edge Recovery Strategy}
\STATE $\mathcal{S}_{\text{cand}} \leftarrow$ Rank non-existing edges in $\mathbf{A}_R$ by scores in $\widetilde{\mathbf{A}}$
\STATE $P \leftarrow \max(\lfloor (p+q)|\mathcal{E}| \rfloor, 0)$
\STATE $\mathbf{A}_P \leftarrow$ Top-$P$ edges from $\mathcal{S}_{\text{cand}}$
\vspace{0.2cm}

\STATE \textbf{Phase 3: Final Graph Construction}
\STATE $\mathbf{A}_E \leftarrow \mathbf{A} - \Delta\mathbf{A} + \alpha \cdot \mathbf{A}_P$ \hfill \COMMENT{\textit{Eq. \ref{eq:final_adjacency}}}
\RETURN $\mathbf{A}_E$
\end{algorithmic}
\end{algorithm}

\section{EXPERIMENTS}\label{secExperiment}

\subsection{Experimental Setup}

\subsubsection{Datasets}
We evaluate our proposed GADPN model on six real-world benchmark datasets, summarized in Table~\ref{tab:dataset}. The datasets encompass diverse domains and structural properties:
\begin{itemize}
    \item \textbf{Citation Networks} \cite{DBLP:conf/iclr/KipfW17}: Cora, Citeseer, and Pubmed are standard citation network datasets where nodes represent academic publications, edges denote citation relationships, node features are bag-of-words representations of the paper abstracts, and labels correspond to the research topic of each paper.
    
    \item \textbf{Wikipedia Networks} \cite{Pei2020Geom-GCN:}: Chameleon and Squirrel are page-page networks extracted from specific Wikipedia topics. Nodes represent web pages, edges represent mutual hyperlinks, node features consist of informative nouns extracted from the pages, and labels indicate the level of web traffic received by each page.
    
    \item \textbf{Actor Co-occurrence Network} \cite{Pei2020Geom-GCN:}: The Actor dataset represents a co-occurrence network of actors, where nodes correspond to actors and edges denote co-occurrence in the same Wikipedia page. Node features are keywords extracted from the actors' Wikipedia pages, and labels represent the types of actors.
\end{itemize}

All datasets are sourced from the PyTorch Geometric library \cite{fey2019fast}. Following the homophily principle, we categorize these benchmarks into two groups: \textit{assortative graphs} (Cora, Citeseer, and Pubmed) and \textit{disassortative graphs} (Chameleon, Squirrel, and Actor). For assortative datasets, we adopt the standard fixed split provided by the library and report the average performance over $10$ repeated runs. For disassortative datasets, we employ random splits and report the average performance over $10$ different train/validation/test partitions to ensure statistical reliability.

\begin{table*}[htbp]
    \centering
    \caption{Statistics of the benchmark datasets used for evaluation.}
    \label{tab:dataset}
    \begin{tabular}{lcccccc}
        \toprule
        \textbf{Dataset} & \textbf{Nodes} ($N$) & \textbf{Edges} ($|\mathcal{E}|$) & \textbf{Classes} & \textbf{Features} ($d$) & \textbf{Train} & \textbf{Val / Test} \\
        \midrule
        Cora & $2{,}708$ & $5{,}429$ & $7$ & $1{,}433$ & $140$ & $500$ / $1{,}000$ \\
        Citeseer & $3{,}327$ & $4{,}732$ & $6$ & $3{,}703$ & $120$ & $500$ / $1{,}000$ \\
        Pubmed & $19{,}717$ & $44{,}338$ & $3$ & $500$ & $60$ & $500$ / $1{,}000$ \\
        \midrule
        Chameleon & $2{,}277$ & $36{,}101$ & $5$ & $2{,}325$ & $1{,}092$ & $729$ / $456$ \\
        Squirrel & $5{,}201$ & $217{,}073$ & $5$ & $2{,}089$ & $2{,}496$ & $1{,}664$ / $1{,}041$ \\
        Actor & $7{,}600$ & $30{,}019$ & $5$ & $932$ & $3{,}648$ & $2{,}432$ / $1{,}520$ \\
        \bottomrule
    \end{tabular}
\end{table*}

\subsubsection{Baselines}
We compare GADPN with three categories of representative GNN models:
\begin{itemize}
    \item \textbf{Spectral-based methods:} GCN \cite{DBLP:conf/iclr/KipfW17} and SGC \cite{wu2019simplifying}.
    \item \textbf{Spatial-based methods:} GraphSAGE \cite{10.5555/3294771.3294869}, GAT \cite{velickovic2018graph}, AGNN \cite{k.2018attentionbased}, APPNP \cite{gasteiger2018combining}, FAGCN \cite{bo2021beyond}, and GATv2 \cite{brody2022how}.
    \item \textbf{Graph structure learning method:} GEN \cite{10.1145/3442381.3449952}.
\end{itemize}

Brief descriptions of each baseline are provided below:
\begin{itemize}
    \item \textbf{GCN}: A semi-supervised graph convolutional network that learns node representations by aggregating features from neighboring nodes.
    \item \textbf{SGC}: A simplified GCN that removes non-linear activations and collapses weight matrices into a linear model while retaining expressive power.
    \item \textbf{GraphSAGE}: An inductive framework that generates node embeddings by sampling and aggregating features from local neighborhoods, enabling generalization to unseen nodes.
    \item \textbf{GAT}: Utilizes attention mechanisms to adaptively weight the importance of different neighbors during feature aggregation.
    \item \textbf{AGNN}: Learns node representations through adaptive information propagation using a self-attention mechanism to weight node influences.
    \item \textbf{APPNP}: Combines neural network predictions with personalized PageRank propagation to iteratively refine node embeddings.
    \item \textbf{FAGCN}: Employs a self-gating mechanism to adaptively integrate different signals during message passing.
    \item \textbf{GATv2}: Improves upon GAT with a more expressive and dynamic attention mechanism, enhancing performance on various graph tasks.
    \item \textbf{GEN}: A graph structure learning model that uses Bayesian inference to jointly optimize GNN parameters and the graph structure.
\end{itemize}

\subsubsection{Implementation Details}
All experiments are conducted on a Linux server equipped with an NVIDIA GeForce RTX $3090$ GPU and an AMD EPYC $7282$ CPU. We use Python $3.12.9$ and PyTorch $2.5.1$.

For baseline implementations, we adopt GCN, SGC, GraphSAGE, GAT, AGNN, APPNP, and GATv2 from the PyTorch Geometric library \cite{fey2019fast}. For FAGCN and GEN, we use the source codes provided by the original authors. All baselines are initialized with the hyperparameters recommended in their respective papers and are further carefully tuned to achieve optimal performance.

For our proposed GADPN, we employ two classic two-layer GNN models, GCN and GraphSAGE, as backbones, both with a fixed hidden dimension of $16$. We use the Adam optimizer with a learning rate ranging from $0.01$ to $0.07$ and a weight decay of $5 \times 10^{-4}$. A dropout rate of $0.5$ and ReLU activation are applied to prevent overfitting. The perturbation parameters are set as follows: $p \in \{0.002, 0.003, \dots, 0.01\}$, $q \in \{-0.002, 0, \dots, 0.04\}$, and $\alpha \in \{0.1, 0.2, \dots, 1\}$.

\begin{table*}[htbp]
    \centering
    \caption{Node classification results (\%). Bold indicates the best performance; underline indicates the runner-up.}
    \label{tab:results}
    \begin{tabular}{l|ccc|ccc}
        \toprule
        \textbf{Datasets} & \textbf{Cora} & \textbf{Citeseer} & \textbf{Pubmed} & \textbf{Chameleon} & \textbf{Squirrel} & \textbf{Actor} \\
        \midrule
        \textbf{GCN}  & $81.0$ & $71.1$ & $77.5$ & $34.4$ & $27.7$ & $28.9$ \\
        \textbf{SGC}  & $80.0$ & $71.3$ & $77.0$ & $34.6$ & $27.8$ & $26.6$ \\
        \midrule
        \textbf{GraphSAGE}  & $80.0$ & $69.9$ & $77.1$ & $46.3$ & $31.4$ & $\underline{33.8}$ \\
        \textbf{GAT}  & $81.7$ & $70.1$ & $76.8$ & $44.1$ & $27.2$ & $28.5$ \\
        \textbf{AGNN}  & $81.2$ & $69.4$ & $78.9$ & $43.4$ & $27.4$ & $32.2$ \\
        \textbf{APPNP} & $\underline{82.6}$ & $71.1$ & $\mathbf{79.9}$ & $39.3$ & $28.5$ & $28.2$ \\
        \textbf{FAGCN}  & $82.0$ & $67.0$ & $77.8$ & $39.3$ & $27.0$ & $33.6$ \\
        \textbf{GATv2}  & $81.5$ & $70.6$ & $77.5$ & $42.6$ & $27.8$ & $28.1$ \\
        \midrule
        \textbf{GEN}  & $81.2$ & $70.8$ & $77.0$ & $57.5$ & $\underline{36.8}$ & $32.8$ \\
        \midrule
        \textbf{GADPN(GCN)}  & $\mathbf{83.2}$ & $\mathbf{72.8}$ & $\underline{79.1}$ & $\underline{60.4}$ & $34.8$ & $29.7$ \\
        \textbf{GADPN(SAGE)}  & $81.7$ & $\underline{71.5}$ & $78.9$ & $\mathbf{61.1}$ & $\mathbf{39.6}$ & $\mathbf{36.4}$ \\
        \bottomrule
    \end{tabular}
\end{table*}

\begin{figure*}[htbp]
    \centering
    \begin{subfigure}{0.325\linewidth}
        \centering
        \includegraphics[width=0.9\linewidth]{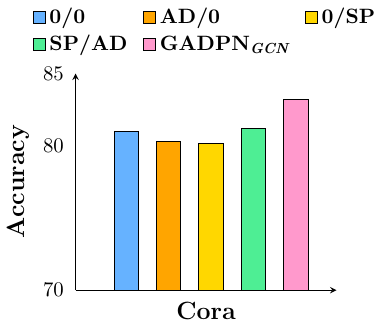}
    \end{subfigure}%
    \begin{subfigure}{0.325\linewidth}
        \centering
        \includegraphics[width=0.9\linewidth]{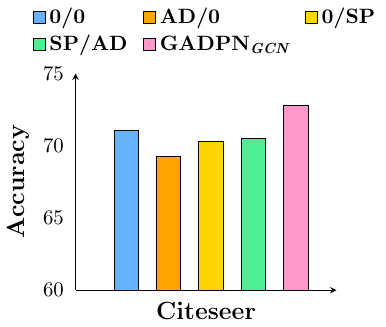}
    \end{subfigure}%
    \begin{subfigure}{0.325\linewidth}
        \centering
        \includegraphics[width=0.9\linewidth]{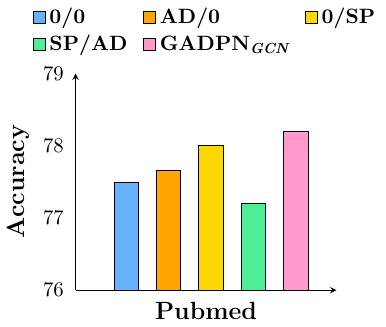}
    \end{subfigure}%
    
    \begin{subfigure}{0.325\linewidth}
        \centering
        \includegraphics[width=0.9\linewidth]{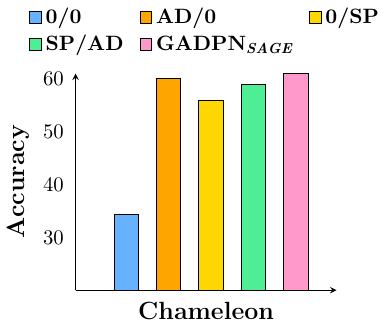}
    \end{subfigure}%
    \begin{subfigure}{0.325\linewidth}
        \centering
        \includegraphics[width=0.9\linewidth]{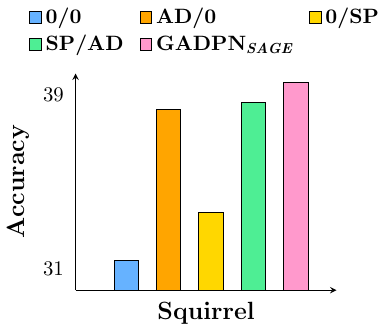}
    \end{subfigure}%
    \begin{subfigure}{0.325\linewidth}
        \centering
        \includegraphics[width=0.9\linewidth]{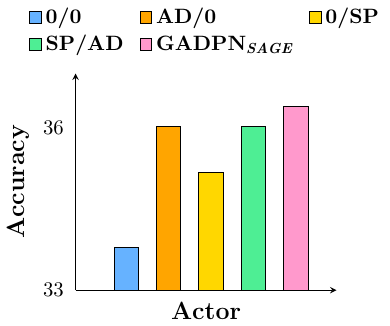}
    \end{subfigure}%
    \caption{Performance comparison (\%) of GADPN and its variants across all datasets.}
    \label{fig:variants}
\end{figure*}

\subsection{Node Classification}
\label{subsec:node_classification}

We evaluate the semi-supervised node classification performance of GADPN against all baseline methods. For each method, we conduct $5$ independent runs with identical hyperparameters and dataset splits, reporting the average accuracy in Table~\ref{tab:results}. The analysis of these results yields the following key observations:

\begin{itemize}
    \item \textbf{Overall Superior Performance:} GADPN achieves the best performance on five out of the six benchmark datasets (Cora, Citeseer, Chameleon, Squirrel, and Actor), demonstrating its effectiveness across diverse graph types. Notably, its advantage is most pronounced on disassortative graphs. On the Pubmed dataset, while GADPN does not attain the top result, it remains highly competitive, securing a performance level close to the best method.

    \item \textbf{Significant Enhancement over Backbones:} GADPN delivers substantial performance gains over its base GNN models. For instance, it achieves a maximum relative improvement of $26\%$ over the standard GCN backbone on the Chameleon dataset. This consistent and significant superiority underscores the efficacy of GADPN's graph structure enhancement module in refining the input topology for downstream tasks.

    \item \textbf{Backbone-specific Trends:} An interesting pattern emerges regarding the choice of backbone network within GADPN. GADPN(GCN) tends to yield superior results on assortative citation networks, whereas GADPN(SAGE) generally excels on the more challenging, disassortative Wikipedia and actor networks. This suggests that the synergy between the enhancement strategy and the aggregation mechanism of the backbone can be dataset-dependent.

    \item \textbf{Simplicity and Efficiency:} Compared to other graph structure learning methods, such as the Bayesian-based GEN, GADPN demonstrates that a principled matrix factorization approach, coupled with adaptive perturbation, can achieve state-of-the-art performance through a relatively simple and efficient pipeline.
\end{itemize}

\subsection{Ablation Study}
\label{subsec:ablation_study}

To verify the effectiveness and necessity of the two core components in GADPN-adaptive denoising (AD) and generalized structural perturbation (SP)-we conduct a comprehensive ablation study. We compare the full GADPN model against three ablated variants, each with a specific component disabled or modified, as well as the original backbone GNN. For assortative datasets (Cora, Citeseer, Pubmed), we use GCN as the backbone; for disassortative datasets (Chameleon, Squirrel, Actor), we use GraphSAGE. The variants are defined as follows:

\begin{itemize}
    \item \textbf{Backbone (0/0):} The base GNN model (GCN or GraphSAGE) without any graph enhancement.
    \item \textbf{AD/0:} GADPN with the structural perturbation module removed, utilizing only adaptive denoising.
    \item \textbf{0/SP:} GADPN with the adaptive denoising module removed, utilizing only structural perturbation on the original graph.
    \item \textbf{SP/AD:} A variant where the order of operations is reversed; structural perturbation is applied first, followed by adaptive denoising.
\end{itemize}

The performance of all variants across the six datasets is illustrated in Fig. \ref{fig:variants}. Our analysis leads to the following conclusions:

\begin{enumerate}
    \item \textbf{Synergy of Components:} The full GADPN model consistently outperforms all ablated variants and the plain backbone across all datasets. This demonstrates that the two components are not merely additive but work synergistically to achieve superior graph enhancement.

    \item \textbf{Necessity of Both Components on Assortative Graphs:} On the assortative Cora and Citeseer datasets, the backbone model alone achieves better performance than either the AD/0 or 0/SP variant. This indicates that on these relatively clean, homophilous graphs, applying only one component in isolation can disrupt the inherent structure and degrade performance. Both components are required to constructively refine the graph.

    \item \textbf{Critical Role on Disassortative Graphs:} In contrast, on the three disassortative datasets (Chameleon, Squirrel, Actor), all three variants (AD/0, 0/SP, and SP/AD) significantly outperform the backbone model. This verifies that both adaptive denoising and structural perturbation are critically important for handling noisy, heterophilous graph structures, where the original topology is less reliable for message passing.

    \item \textbf{Importance of Execution Order:} The SP/AD variant, which reverses the component order, underperforms compared to the standard GADPN (AD then SP). This result empirically validates our design intuition: it is more effective to first \textit{denoise} the graph to obtain a cleaner signal and then apply \textit{perturbation} to enhance its robustness, rather than perturbing a noisy graph first. The proposed order provides a more stable foundation for the perturbation-recovery process.
\end{enumerate}

In summary, the ablation study confirms that both adaptive denoising and generalized structural perturbation are indispensable components of GADPN. Their specific sequential integration is crucial for unlocking the full potential of graph structure learning across diverse network types.

\begin{figure*}[htbp]
    \centering
    \begin{subfigure}{0.2\linewidth}
        \centering
        \includegraphics[width=\linewidth]{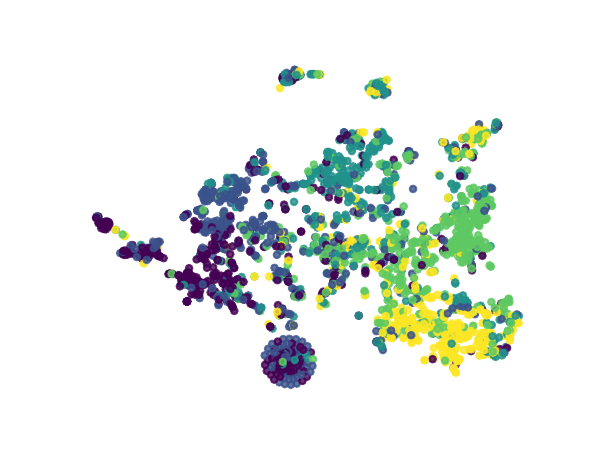}
        \caption{GCN}
    \end{subfigure}%
    \hspace{-0.3em}%
    \begin{subfigure}{0.2\linewidth}
        \centering
        \includegraphics[width=\linewidth]{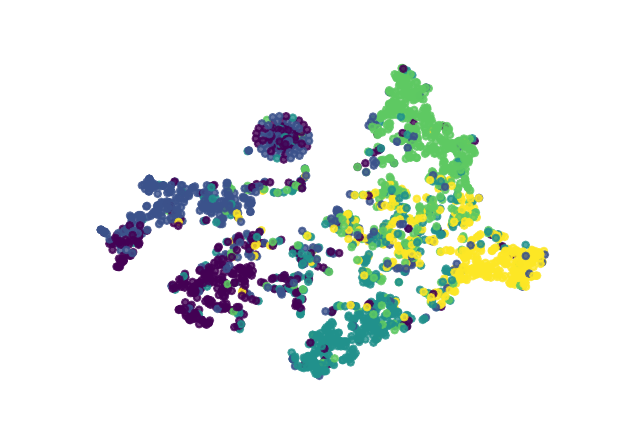}
        \caption{GraphSAGE}
    \end{subfigure}%
    \hspace{-0.3em}%
    \begin{subfigure}{0.2\linewidth}
        \centering
        \includegraphics[width=\linewidth]{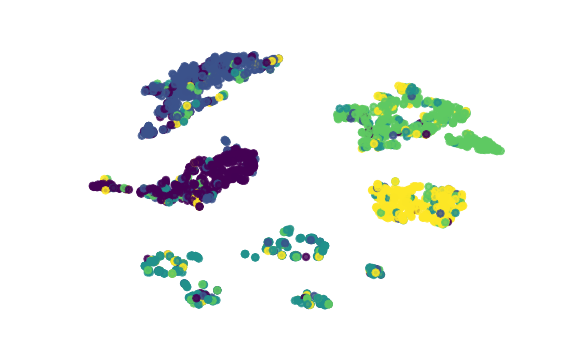}
        \caption{GEN}
    \end{subfigure}%
    \hspace{-0.3em}%
    \begin{subfigure}{0.2\linewidth}
        \centering
        \includegraphics[width=\linewidth]{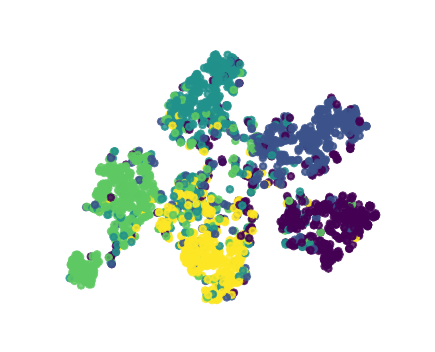}
        \caption{GADPN}
    \end{subfigure}
    \caption{Visualization of the learned node embeddings on the Chameleon dataset.}
    \label{fig:vis}
\end{figure*}

\subsection{Embedding Visualization}
\label{subsec:visualization}

To provide an intuitive assessment of the quality of the learned representations and further validate the effectiveness of our proposed model, we conduct a qualitative embedding visualization on the Chameleon dataset. Specifically, we extract the output embeddings from the last layer (prior to the softmax activation) of each model-GADPN, GCN, GraphSAGE, and the graph structure learning baseline GEN. We then reduce these high-dimensional embeddings to a two-dimensional space using t-SNE \cite{van2008visualizing} and visualize the resulting projections of the test set nodes, coloring each point according to its ground-truth class label.

The visualization results are presented in Fig. \ref{fig:vis}. We observe distinct patterns in the embedding structures learned by different models:
\begin{itemize}
    \item \textbf{GCN \& GraphSAGE:} The embeddings produced by the standard GCN and GraphSAGE models exhibit significant overlap and mixing among different classes. The points are scattered across the entire region without forming well-separated clusters, indicating that these models struggle to learn discriminative representations on this heterophilous graph.

    \item \textbf{GEN:} The baseline graph structure learning model, GEN, performs somewhat better, showing a degree of separation between some classes. However, the clusters remain relatively diffuse, and inter-class boundaries are not clearly defined.

    \item \textbf{GADPN:} In contrast, GADPN learns embeddings that form compact, well-separated clusters corresponding to the different classes. The visualization demonstrates high intra-class similarity and clear inter-class boundaries, which correlates with its superior classification accuracy. This suggests that GADPN's graph enhancement process effectively transforms the input graph into a structure that facilitates the learning of more discriminative node representations.
\end{itemize}

These visual results provide compelling qualitative evidence that GADPN's adaptive denoising and structural perturbation mechanisms successfully refine the graph topology, enabling the GNN backbone to capture more meaningful and separable node features for classification.

\begin{table*}[htbp]
    \centering
    \caption{Analysis of the adaptive denoising module: optimal rank $k^*$ selected via Bayesian optimization.}
    \label{tab:rank_analysis}
    \begin{tabular}{l|ccc|ccc}
        \toprule
        \textbf{Metric} & \textbf{Cora} & \textbf{Citeseer} & \textbf{Pubmed} & \textbf{Chameleon} & \textbf{Squirrel} & \textbf{Actor} \\
        \midrule
        Optimal rank, $k^*$ & $1{,}950$ & $2{,}519$ & $14{,}173$ & $1{,}588$ & $3{,}210$ & $3{,}076$ \\
        Ratio, $k^*/N$ & $72.01\%$ & $75.71\%$ & $71.95\%$ & $69.77\%$ & $61.72\%$ & $40.48\%$ \\
        \bottomrule
    \end{tabular}
\end{table*}

\subsection{Analysis of Adaptive Denoising}
\label{subsec:rank_analysis}

To gain deeper insight into the adaptive denoising mechanism, we analyze the optimal rank $k^*$ selected by the Bayesian optimization module for GADPN (SAGE) across all datasets. As detailed in Table~\ref{tab:rank_analysis}, we report both the absolute optimal rank $k^*$ and its relative proportion $k^*/N$, where $N$ is the total number of nodes. This proportion quantifies the percentage of the singular value spectrum retained during the low-rank approximation.

The experimental procedure for determining $k^*$ was as follows: For the assortative datasets (Cora, Citeseer, Pubmed), we conducted $5$ independent runs with $50$ Bayesian optimization iterations each and averaged the resulting $k^*$ values. For the disassortative datasets (Chameleon, Squirrel, Actor), we performed the same optimization across $10$ random data splits and computed the average.

The results reveal a clear and interpretable pattern:
\begin{itemize}
    \item For the assortative datasets (Cora, Citeseer, Pubmed), the $k^*/N$ ratios are consistently high, exceeding $70\%$. This indicates that on graphs with strong homophily, the denoising process preserves a majority of the original structural information. The underlying topology is already relatively reliable, and the primary role of adaptive denoising is to perform mild smoothing rather than aggressive filtering.

    \item For the disassortative datasets (Chameleon, Squirrel, Actor), the $k^*/N$ ratios are lower, all below $70\%$. Notably, for the Actor dataset, the ratio drops to approximately $40.4\%$. During the optimization, the algorithm even selected $k^* = 1$ in two instances, suggesting that the original graph structure is highly noisy or misleading for this task. This behavior aligns with our expectation that disassortative graphs often contain more spurious connections or complex, non-homophilous patterns that necessitate a more aggressive low-rank approximation to extract the useful signal.
\end{itemize}

These findings demonstrate that the adaptive denoising module functions effectively across both homophilous and heterophilous contexts. Crucially, it automatically adjusts its denoising strength in a data-driven manner: it retains more information for assortative graphs while performing more aggressive compression for disassortative ones. This adaptability is a key factor in GADPN's robust performance across diverse graph types.

\begin{figure}[htbp]
    \centering
    \begin{subfigure}{0.45\linewidth}
        \centering
        \includegraphics[width=\linewidth]{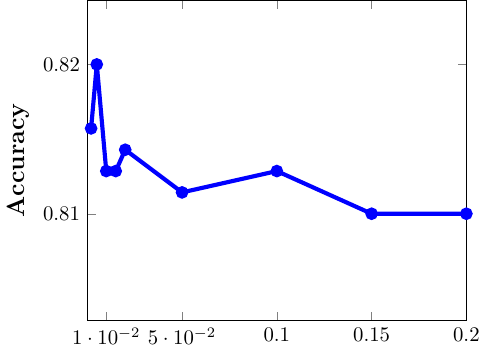}
        \caption{Cora}
    \end{subfigure}
    \hfill
    \begin{subfigure}{0.45\linewidth}
        \centering
        \includegraphics[width=\linewidth]{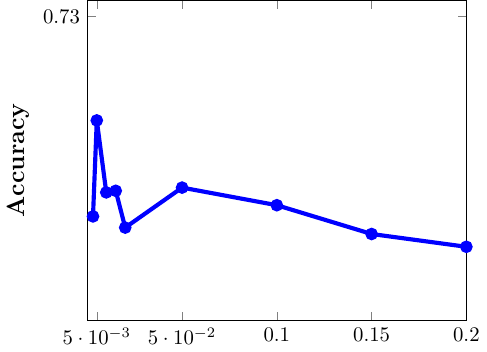}
        \caption{Citeseer}
    \end{subfigure}

    \vspace{0.5cm}

    \begin{subfigure}{0.45\linewidth}
        \centering
        \includegraphics[width=\linewidth]{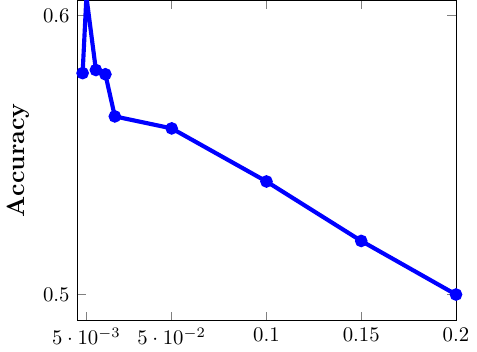}
        \caption{Chameleon}
    \end{subfigure}
    \hfill
    \begin{subfigure}{0.45\linewidth}
        \centering
        \includegraphics[width=\linewidth]{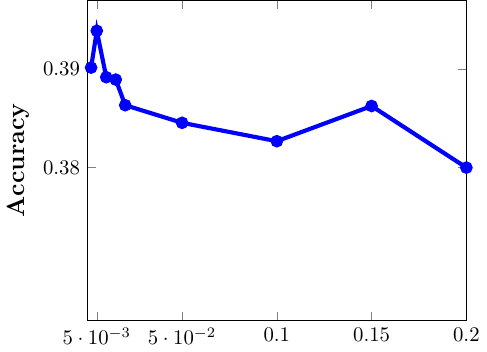}
        \caption{Squirrel}
    \end{subfigure}

    \caption{Performance sensitivity analysis with respect to the base perturbation ratio $p$.}
    \label{fig:p_analysis}
\end{figure}

\begin{figure}[htbp]
    \centering
    \begin{subfigure}{0.45\linewidth}
        \centering
        \includegraphics[width=\linewidth]{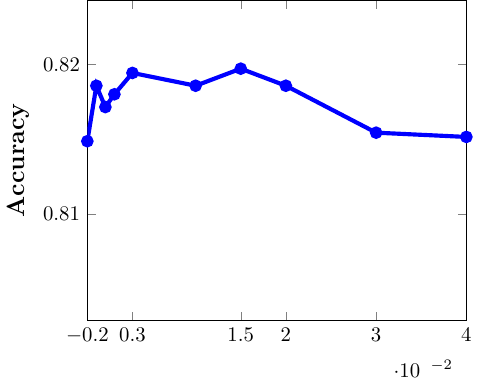}
        \caption{Cora}
    \end{subfigure}
    \hfill
    \begin{subfigure}{0.45\linewidth}
        \centering
        \includegraphics[width=\linewidth]{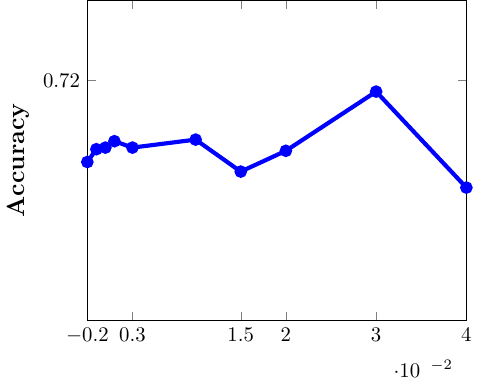}
        \caption{Citeseer}
    \end{subfigure}

    \vspace{0.5cm}

    \begin{subfigure}{0.45\linewidth}
        \centering
        \includegraphics[width=\linewidth]{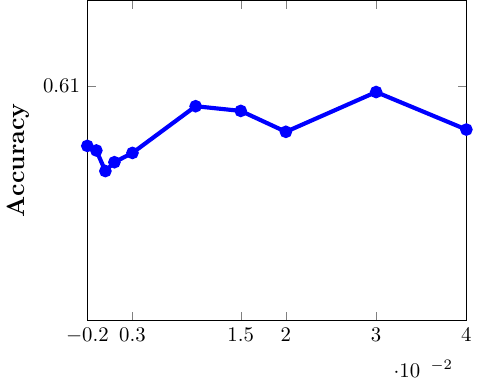}
        \caption{Chameleon}
    \end{subfigure}
    \hfill
    \begin{subfigure}{0.45\linewidth}
        \centering
        \includegraphics[width=\linewidth]{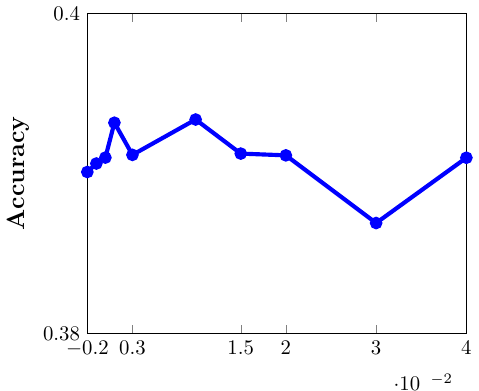}
        \caption{Squirrel}
    \end{subfigure}

    \caption{Performance sensitivity analysis with respect to the recovery adjustment parameter $q$.}
    \label{fig:q_analysis}
\end{figure}

\subsection{Parameter Sensitivity Analysis}
\label{subsec:parameter_study}

In this section, we conduct a detailed sensitivity analysis of GADPN's key hyperparameters: the base perturbation ratio $p$, the recovery adjustment parameter $q$, and the recovered-edge weight $\alpha$. We investigate their effects on the classification performance using GraphSAGE as the backbone model across four representative datasets (Cora, Citeseer, Chameleon, and Squirrel). The results are visualized in Figures \ref{fig:p_analysis}, \ref{fig:q_analysis}, and \ref{fig:alpha_analysis}.

\subsubsection*{Analysis of the Base Perturbation Ratio $p$}
The parameter $p$ controls the fraction of edges temporarily removed for the structural perturbation. We vary $p$ from $0.0$ to $0.2$. As shown in Figure \ref{fig:p_analysis}, model performance generally exhibits a concave pattern: it first improves with increasing $p$, reaches an optimal point, and then declines. This trend can be explained by the dual role of perturbation. A moderate $p$ effectively tests and reinforces the structural consistency of the denoised graph, acting as a form of regularization that improves generalization. However, an excessively large $p$ removes too many edges, causing significant deviation from the original stable structure and degrading the signal available for the recovery process, ultimately harming performance. This demonstrates the importance of selecting an appropriate perturbation strength.

\subsubsection*{Analysis of the Recovery Adjustment $q$}
The parameter $q$ offers flexibility in the edge recovery strategy by adjusting the number of edges $P = \lfloor (p+q)|\mathcal{E}| \rfloor$ to be recovered from the perturbed matrix. We vary $q$ from $-0.002$ to $0.04$. Performance trends across the four datasets, shown in Figure \ref{fig:q_analysis}, indicate a general preference for a positive $q$ value. A $q > 0$ corresponds to \textbf{exploratory recovery}, where the model is allowed to recover more potential edges than were removed. This often proves beneficial as it can reintroduce genuinely important connections that were missing in the observed graph. A $q \leq 0$ results in conservative or reduced recovery, which may fail to fully exploit the model's capacity to correct the graph structure. The optimal $q$ is dataset-dependent, reflecting the varying density of missing true links in different networks.

\subsubsection*{Analysis of the Recovered-Edge Weight $\alpha$}
\begin{figure}[htbp]
    \centering
    \begin{subfigure}{0.45\linewidth}
        \centering
        \includegraphics[width=\linewidth]{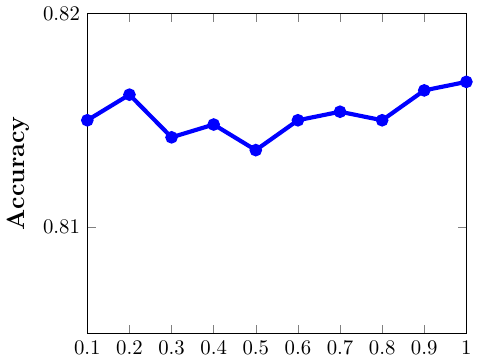}
        \caption{Cora}
    \end{subfigure}
    \hfill
    \begin{subfigure}{0.45\linewidth}
        \centering
        \includegraphics[width=\linewidth]{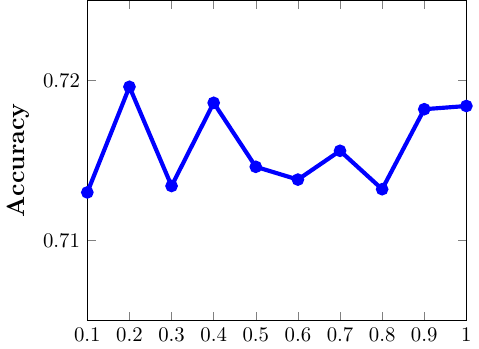}
        \caption{Citeseer}
    \end{subfigure}

    \vspace{0.5cm}

    \begin{subfigure}{0.45\linewidth}
        \centering
        \includegraphics[width=\linewidth]{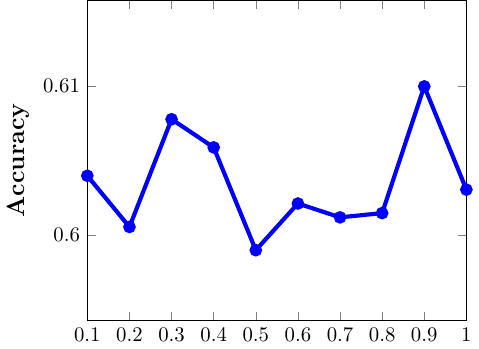}
        \caption{Chameleon}
    \end{subfigure}
    \hfill
    \begin{subfigure}{0.45\linewidth}
        \centering
        \includegraphics[width=\linewidth]{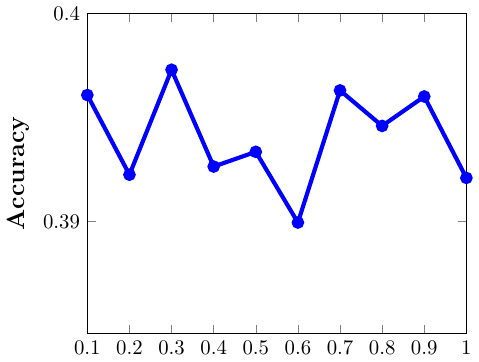}
        \caption{Squirrel}
    \end{subfigure}

    \caption{Sensitivity analysis with respect to the recovered-edge weight $\alpha$.}
    \label{fig:alpha_analysis}
\end{figure}

The parameter $\alpha$ governs the influence of the recovered edges $\mathbf{A}_P$ on the final enhanced adjacency matrix $\mathbf{A}_E$, as defined in Equation~\eqref{eq:final_adjacency}. We analyze its effect by varying $\alpha$ from $0.1$ to $1.0$. The results in Figure \ref{fig:alpha_analysis} reveal differing sensitivity patterns. Performance on the assortative Cora and Citeseer datasets shows relatively small fluctuations, indicating that these graphs are less sensitive to the precise weighting of recovered links, likely due to their already high-quality structure. In contrast, performance on the disassortative Chameleon and Squirrel datasets fluctuates more significantly with $\alpha$. This higher sensitivity suggests that for noisy, heterophilous graphs, carefully calibrating the contribution of newly recovered edges is crucial for balancing the injection of new information against the introduction of potential noise.

In summary, the parameter study confirms that GADPN's components are controllable and interpretable. The optimal configuration is data-dependent, yet the model demonstrates stable performance within reasonable parameter ranges, showcasing its robustness.

\section{Conclusion}
\label{secConclusion}

In this paper, we have presented GADPN, a novel and lightweight graph structure learning framework that synergistically integrates adaptive denoising with generalized structural perturbation to enhance graph neural networks. The core of our contribution lies in two key advancements: first, we introduce a Bayesian-optimized, adaptive low-rank approximation that automatically tailors the denoising strength to the inherent homophily and noise level of each specific graph. Second, we generalize the structural perturbation method from symmetric to arbitrary graphs by formulating it through singular value decomposition, thereby extending its applicability to directed and heterogeneous networks.

Extensive experiments on six benchmark datasets demonstrate that GADPN consistently outperforms a wide range of state-of-the-art GNNs and graph structure learning methods. Notably, it achieves particularly significant performance gains on challenging disassortative and noisy graphs, where traditional GNNs often struggle. Ablation studies confirm the necessity and synergy of both core components, while parameter analyses reveal the model's robustness and interpretability.

Looking forward, several promising directions remain open for exploration. The framework could be extended to dynamic graph settings to handle evolving structures. Furthermore, investigating alternative or more efficient matrix factorization techniques, as well as integrating causal reasoning to mitigate confounding effects during augmentation, could further enhance the model's generality and robustness.

 





\bibliographystyle{IEEEtran} 
\bibliography{references}     

\end{document}